\documentclass[journal,twoside,print]{ieeecolor}
\usepackage{nologo}
\usepackage{cite}
\usepackage{amsmath,amssymb,amsfonts}
\usepackage{algorithmic}
\usepackage{graphicx}
\usepackage{textcomp}
\usepackage[T1]{fontenc}
\usepackage{latexsym}
\usepackage[misc,geometry]{ifsym} 
\usepackage{bbm}
\usepackage{float}
\usepackage{subcaption}
\usepackage{booktabs} 

\usepackage{hyperref}
\usepackage{amssymb}
\usepackage{amsmath}
\usepackage{gensymb}
\usepackage{algorithm}
\usepackage[font=small,labelfont=bf]{caption}
\usepackage{color}
\usepackage{colortbl}
\usepackage{multirow}

\makeatother
\DeclareMathOperator*{\argmax}{argmax}

\usepackage{xr}
\makeatletter
\newcommand*{\addFileDependency}[1]{
  \typeout{(#1)}
  \@addtofilelist{#1}
  \IfFileExists{#1}{}{\typeout{No file #1.}}
}
\makeatother
\newcommand*{\myexternaldocument}[1]{%
    \externaldocument{#1}%
    \addFileDependency{#1.tex}%
    \addFileDependency{#1.aux}%
}
\myexternaldocument{supplement}

\def\BibTeX{{\rm B\kern-.05em{\sc i\kern-.025em b}\kern-.08em
    T\kern-.1667em\lower.7ex\hbox{E}\kern-.125emX}}
\begin{document}

\title{An efficient deep neural network to find small objects in large 3D images}

\author{Jungkyu Park,
Jakub Ch\l{}\k{e}dowski, 
Stanisław Jastrzębski,
Jan Witowski,
Yanqi Xu,
Linda Du,\\ 
Sushma Gaddam, 
Eric Kim, 
Alana Lewin, 
Ujas Parikh, 
Anastasia Plaunova,
Sardius Chen, \\
Alexandra Millet, 
James Park, 
Kristine Pysarenko, 
Shalin Patel,
Julia Goldberg, 
Melanie Wegener, \\
Linda Moy, 
Laura Heacock, 
Beatriu Reig,
and Krzysztof J. Geras
\thanks{This work was supported in part by grants from the National Institutes of Health (P41EB017183, R21CA225175), the National Science Foundation (1922658), the Gordon and Betty Moore Foundation (9683), Polish National Science Center (2021/41/N/ST6/02596), and the Polish National Agency for Academic Exchange (PPN/IWA/2019/1/00114/U/00001).}
\thanks{J. Park, S. Jastrzębski, J. Witowski, L. Du, S. Gaddam, E. Kim, A. Lewin, U. Parikh, A. Plaunova, S. Chen, A. Millet, J. Park, K. Pysarenko, S. Patel, J. Goldberg, M. Wegener, L. Moy, L. Heacock, B. Reig, K. J. Geras are with the Department of Radiology, NYU Langone Health, New York, NY 10016 USA (email: jungkyu.park@nyulangone.org; staszek.jastrzebski@gmail.com; jwitos@gmail.com; linda.h.du@gmail.com; Sushma.Gaddam@nyulangone.org; Eric.Kim@nyulangone.org; Alana.Amarosa@nyulangone.org; Ujas.Parikh@ucsf.edu; Anastasia.Plaunova@nyulangone.org; sardiuschen@gmail.com; amillet2016@gmail.com; James.Park2@nyulangone.org; Kristine.Pysarenko@nyulangone.org; Shalin.Patel2@nyulangone.org; Julia.Goldberg@nyulangone.org; melanie.wegener@nyulangone.org; Linda.Moy@nyulangone.org; Laura.Heacock@nyulangone.org; beatriu.reig@nyulangone.org; k.j.geras@nyu.edu).}
\thanks{J. Ch\l{}\k{e}dowski is with the Faculty of Mathematics and Computer Science, Jagiellonian University, Kraków 30-348 Poland (email: jakub.chledowski@gmail.com).}
\thanks{Y. Xu is with the Center for Data Science, New York University, New York, NY 10011 USA (email: yx2105@nyu.edu).}
\thanks{This work has been submitted to the IEEE for possible publication. Copyright may be transferred without notice, after which this version may no longer be accessible.}
}

\maketitle

\begin{abstract}
3D imaging enables accurate diagnosis by providing spatial information about organ anatomy. 
However, using 3D images to train AI models is computationally challenging because they consist of 10x or 100x more pixels than their 2D counterparts. 
To be trained with high-resolution 3D images, convolutional neural networks resort to downsampling them or projecting them to 2D. 
We propose an effective alternative, a neural network that enables efficient classification of full-resolution 3D medical images.
Compared to off-the-shelf convolutional neural networks, our network, 3D Globally-Aware Multiple Instance Classifier (3D-GMIC), uses 77.98\%-90.05\% less GPU memory and 91.23\%-96.02\% less computation.
While it is trained only with image-level labels, without segmentation labels, it explains its predictions by providing pixel-level saliency maps.
On a dataset collected at NYU Langone Health, including 85,526 patients with full-field 2D mammography (FFDM), synthetic 2D mammography, and 3D mammography, 3D-GMIC achieves an AUC of 0.831 (95\% CI: 0.769-0.887) in classifying breasts with malignant findings using 3D mammography. 
This is comparable to the performance of GMIC on FFDM (0.816, 95\% CI: 0.737-0.878) and synthetic 2D (0.826, 95\% CI: 0.754-0.884), which demonstrates that 3D-GMIC successfully classified large 3D images despite focusing computation on a smaller percentage of its input compared to GMIC.
Therefore, 3D-GMIC identifies and utilizes extremely small regions of interest from 3D images consisting of hundreds of millions of pixels, dramatically reducing associated computational challenges.
3D-GMIC generalizes well to BCS-DBT, an external dataset from Duke University Hospital, achieving an AUC of 0.848 (95\% CI: 0.798-0.896).
\end{abstract}


\section{Introduction}
\label{sec:introduction}
Mammography is the standard medical imaging modality in breast cancer screening programs, which are crucial in the early detection of the leading cause of cancer-related death among women worldwide~\cite{sung2021globocan}.
Digital breast tomosynthesis (DBT) or ``3D mammography'', has been introduced to improve quality of diagnosis in breast cancer screening.
Full-field digital mammography (FFDM) projects the three-dimensional information of breasts to two-dimensional planes, which can obscure lesions with other breast tissues.
Because DBT images are three-dimensional, they more accurately capture the three-dimensional information than FFDM and enable better identification of suspicious findings.
Thus, DBT leads to more cancers being found and a reduction in recall rates~\cite{kopans2014digital, mcdonald2016effectiveness, rafferty2016breast, conant2019association, conant2020five, bahl2020breast}.
However, as DBT images contain 70 slices on average (about 340 million pixels on average), it almost doubles its interpretation time for radiologists compared to FFDM~\cite{aase2019randomized}.
Therefore, there is a need for tools for computer-aided detection and diagnosis to reduce interpretation time for DBT images. 

One cannot naively apply large-capacity neural networks to high-resolution multi-slice DBT images with modern GPUs without running out of memory. 
Attempts to downsample 3D images or project each 3D image into one 2D image lead to losing the benefit of DBT: separation of lesions from nearby structures. 
If pixel-level or lesion-level ground truth labels (segmentations of regions of interest) were available, it is possible to train neural networks by utilizing small subset of large 3D images at a time.
However, collecting these annotations is a time-consuming and mostly manual process, which requires experts (e.g., radiologists) to annotate suspicious areas on dozens of slices.

To address these limitations, we propose 3D Globally-Aware Multiple Instance Classifier (3D-GMIC), a deep neural network which predicts malignancy of lesions and localizes them in 3D images. 
Our architecture extends the Globally-Aware Multiple Instance Classifier (GMIC)~\cite{shen2021interpretable}, a highly effective neural network architecture for large 2D images, to 3D data.
3D-GMIC acts by first determining the most important regions utilizing a low-capacity network, to which a high-capacity network is applied.
3D-GMIC learns to highlight the regions important to predicting image-wise classification labels. 
This enables applying 3D-GMIC to learn from data without segmentation labels.
Importantly, among similar regions in nearby slices, 3D-GMIC selects the region in the most important slice and avoids processing duplicate information.
This enables training deep neural networks with entire DBT images without downsampling or performance degradation.
In addition, we improve the stability of learning in low-data regimes when using backbone networks with ReLU nonlinearity. 

To demonstrate the effectiveness of 3D-GMIC, we compare its performance to models trained on equivalent 2D datasets.
We train and evaluate the models on three different imaging modalities used in breast cancer screening; (1) FFDM: Full-Field Digital Mammography, (2) DBT: Digital Breast Tomosynthesis and (3) synthetic 2D: 2D image generated by combining information from different slices of DBT images using the proprietary C-View\texttrademark \phantom{ } algorithm~\cite{smith2016synthesized}.
We show that our 3D-GMIC model enables end-to-end training with whole DBT images by using 77.98\%-90.05\% less GPU memory and 91.23\%-96.02\% less computation compared to off-the-shelf deep convolutional neural networks.
In addition, we show that 3D-GMIC trained on our internal dataset generalizes well to an external dataset from Duke University Hospital~\cite{buda2021data, buda2020data, clark2013cancer}.
Hence 3D-GMIC may allow other researchers to perform classification and weakly-supervised semantic segmentation with other types of 3D data.
Our model code and model weights are released at \url{https://github.com/nyukat/3D_GMIC}.

\section{Related Work}

Prior works on building deep neural networks for DBT images have at least one of the following four disadvantages.

First, many of them work with synthesized 2D images rather than the entire 3D image.
These images are generated from the DBT images by proprietary algorithms bundled with the scanner~\cite{matthews2020multi} or by third-party methods aggregating information from all slices such as dynamic feature image~\cite{liang2019joint} or maximum intensity projection~\cite{singh2020adaptation}. 
In addition, there exist approaches which create attention-weighted summaries of small subsets of DBT slices, resulting in multiple ``slabs'' that resemble 2D mammography~\cite{tardy2021trainable}.
The resulting images may suffer from the same disadvantage as FFDM: the lesion could be hidden by nearby structures located at different depths.
Therefore, the performances of the model on synthesized 2D images could suffer compared to model on DBT images.
In addition, models trained on synthesized 2D images will not be able to predict which slices in DBT images contain the suspicious findings, limiting the usefulness in assisting the radiologists in reading DBT images.

Second, the studies that use DBT instead of synthetic 2D images still do not utilize the entire image at once in training. 
Some utilize a subset of DBT slices~\cite{zhang2018classification} whereas others process small region of interest (ROI) patches pre-extracted from lesions~\cite{samala2018evolutionary, li2020digital, singh2020adaptation}.
A model utilizing a subset of slices risks missing those potentially most informative to the diagnosis.
A model processing image patches does not learn to utilize the global context of the entire breast. 
 

Third, some of the works require more detailed labels such as bounding-boxes or pixel-level segmentations~\cite{fan2019computer, lai2020dbt, lotter2021robust, buda2021data, samala2018evolutionary, li2020digital, singh2020adaptation}.
While training the model with pixel or lesion-level labels may improve the performance compared to just using image-level labels, the former are more labor-intensive and time-consuming to collect, especially for 3D data. Collecting segmentation or bounding box labels requires experts to perform manual work, while study-level labels can be reliably extracted from hospital information systems. 

Lastly, most of the prior works were evaluated on datasets with small sample sizes~\cite{samala2016mass, lai2020dbt, fotin2016detection, mendel2019transfer} or patient cohorts with specific inclusion criteria.
In addition, they do not perform any external validation of their models using datasets from other hospitals.
This limits drawing reliable conclusions about the potential clinical utility of such models.

More details on the related works on AI for DBT can be found in Bai et al. (2021)~\cite{BAI2021102049}.
For example, Singh et al. (2020)~\cite{singh2020adaptation} achieved an AUC of 0.847, Tardy and Mateus (2021)~\cite{tardy2021trainable} achieved an AUC of 0.73.
Please note that the reported performances of the related works are not directly comparable because they are not evaluated on the same dataset.

In addition, there exist general-purpose computation- and memory-saving techniques such as gradient checkpointing~\cite{chen2016training} and swapping CPU and GPU memory~\cite{matzek2018data, rasley2020deepspeed} which reduce the maximum GPU memory usage to enable training high-capacity neural networks with large 3D images without downsampling. However, these techniques do not reduce the amount of required computation. 
In fact, they add additional overhead for recomputing forward passes, saving and loading gradients, and/or swapping memories between devices.

\section{Dataset}

\subsection{Internal Dataset}
Our retrospective study was approved by the NYU Langone Health Institutional Review Board (study ID\# i18-00712) and was compliant with the Health Insurance Portability and Accountability Act. Informed consent was waived.
To perform this study, we created a dataset consisting of exams containing both FFDM and DBT images.
We refer to this dataset as ``NYU Combo v1''.
The NYU Combo v1 dataset consists of 99,862 exams from 85,526 patients screened between January 2016 and June 2018 at NYU Langone Health. 
Each exam contains the four standard views used in screening mammography (R-CC, L-CC, R-MLO, and L-MLO) for all three imaging modalities (FFDM, DBT, synthetic 2D).
The manufacturer and model name of all images in this dataset are 'HOLOGIC, Inc.' and 'Selenia Dimensions'.
Since 99\% of DBT images have 96 slices or less, we discarded exams that contain DBT images with more than 96 slices.
This is to keep the efficiency of data loading and confine the VRAM usage to a predictable amount during our experiments. 
Note that it is possible to utilize more than 96 slices for both training and inference as long as sufficient VRAM is available.

Only a small number of breasts are associated with pathology labels.
Among the 199,724 breasts in the NYU Combo v1 dataset, 2,579 breasts (1.29\%) had benign findings and 635 breasts (0.32\%) had malignant findings. 
150 (0.08\%) breasts had both benign and malignant findings.
All benign and malignant findings were pathology-proven.
Exams not associated with any pathology report were assigned a negative label, indicating the absence of any biopsied findings.
Note that NYU Combo v1 dataset is about half the size of the dataset used in Wu et al.~\cite{wu2019deep} and Shen et al.~\cite{shen2021interpretable}.

We divided NYU Combo v1 into training, validation, and test sets.
First, we sorted patients in the chronological order of their most recent exams.
We designated the first (i.e., the ones who had their last exam earlier than other patients) 80\% of the patients (68,412) to the training set, the next 10\% of the patients (8,543) to the validation set, and the remaining (most recent) 10\% (8,571) to the test set.
This results in 78,702, 10,266 and 10,894 exams in the training, validation, and the test set respectively.
There are 518, 64, 53 malignant breasts and 2095, 245, 239 benign breasts in the training, validation, test sets respectively.

To acquire lesion annotations on DBT images in the test set, we asked a group of 16 radiologists from NYU Langone Health to annotate the location of biopsied lesions by segmenting them with ITK-SNAP~\cite{yushkevich2016itk}.
We used these labels only for evaluating semantic segmentation performance in the test set. They were not used during training and validation.

In addition, as only a small number of exams have benign or malignant findings, we utilize transfer learning to improve the performance of our models as described by Wu et al. and Shen et al.~\cite{wu2019deep, shen2021interpretable}. 
In this pretraining, we utilized BI-RADS labels extracted from radiology reports, which are the radiologist's assessment of a patient's risk of having breast cancer based on the screening mammography.
We use all BI-RADS scores but group them into 3 classes: (a) incomplete or high suspicion (BI-RADS category 0, 4, 5), (b) normal (BI-RADS category 1), and (c) benign (BI-RADS category 2 and 3).
We utilized both the NYU Breast Cancer Screening Dataset v1.0 (BCSDv1)~\cite{wu2019nyu} and NYU Combo v1 in pretraining a part of our model.
When combining the two datasets, we only selected patients from BCSDv1 who do not overlap with the patients in the validation and test sets of NYU Combo v1 to avoid an information leak.
As a result, we were able to utilize additional 210,389 2D screening mammography exams for pretraining our models with BI-RADS labels.
More details of pretraining and transfer learning are in section \ref{sec:birads}. 

\subsection{External Dataset}
In an additional evaluation of our models, we utilize BCS-DBT, an external dataset from Duke University Hospital~\cite{buda2021data, buda2020data, clark2013cancer}, specifically the subset that was released as the training dataset of the DBTex challenge\footnote{\url{https://spie-aapm-nci-dair.westus2.cloudapp.azure.com/competitions/9}}~\cite{park2021lessons}.
This dataset consists of 19,148 DBT images from 4,838 studies that belong to 4,362 patients.
We utilize all images, including the ones with 96 slices or more.
The manufacturer and model name of all images in this dataset are 'HOLOGIC, Inc.' and 'Selenia Dimensions'.
It contains 87 bounding-box labels for malignant lesions, and 137 bounding-box labels for benign lesions.
Even though this dataset uses the same manufacturer and model as NYU Combo V1 dataset, BCS-DBT still serves as an effective dataset for external validation. 
This is because there are variations in data collection and processing between datasets collected from different hospitals, which could degrade cross-institutional generalization~\cite{zech2018variable}.

\section{Task definition}

We formulate the task of predicting the probability of presence of benign and malignant lesions as a multi-label classification problem. That is, given an image $\mathbf{x} \in \mathbb{R}^{H,W,D}$, our models make probability predictions $\mathbf{p}$ corresponding to the labels $\mathbf{y} = \begin{bmatrix} y^b \\ y^m \end{bmatrix}$ for each image, where $y^b, y^m \in \{0,1\}$ indicate the presence of at least one biopsy-confirmed benign or malignant lesion in $\mathbf{x}$, respectively.

\section{AI architecture}\label{sec:arch}

The proposed architecture, 3D-GMIC, consists of two sub-networks: \textit{the global module} and \textit{the local module} (Fig.~\ref{fig:overall_plot}).

\subsection{High-level overview of 3D-GMIC}\label{sec:overview}

\subsubsection{The global module} 

\begin{figure*}
  \centering
 \includegraphics[width=\textwidth, trim={0 0 4.5cm 0}]{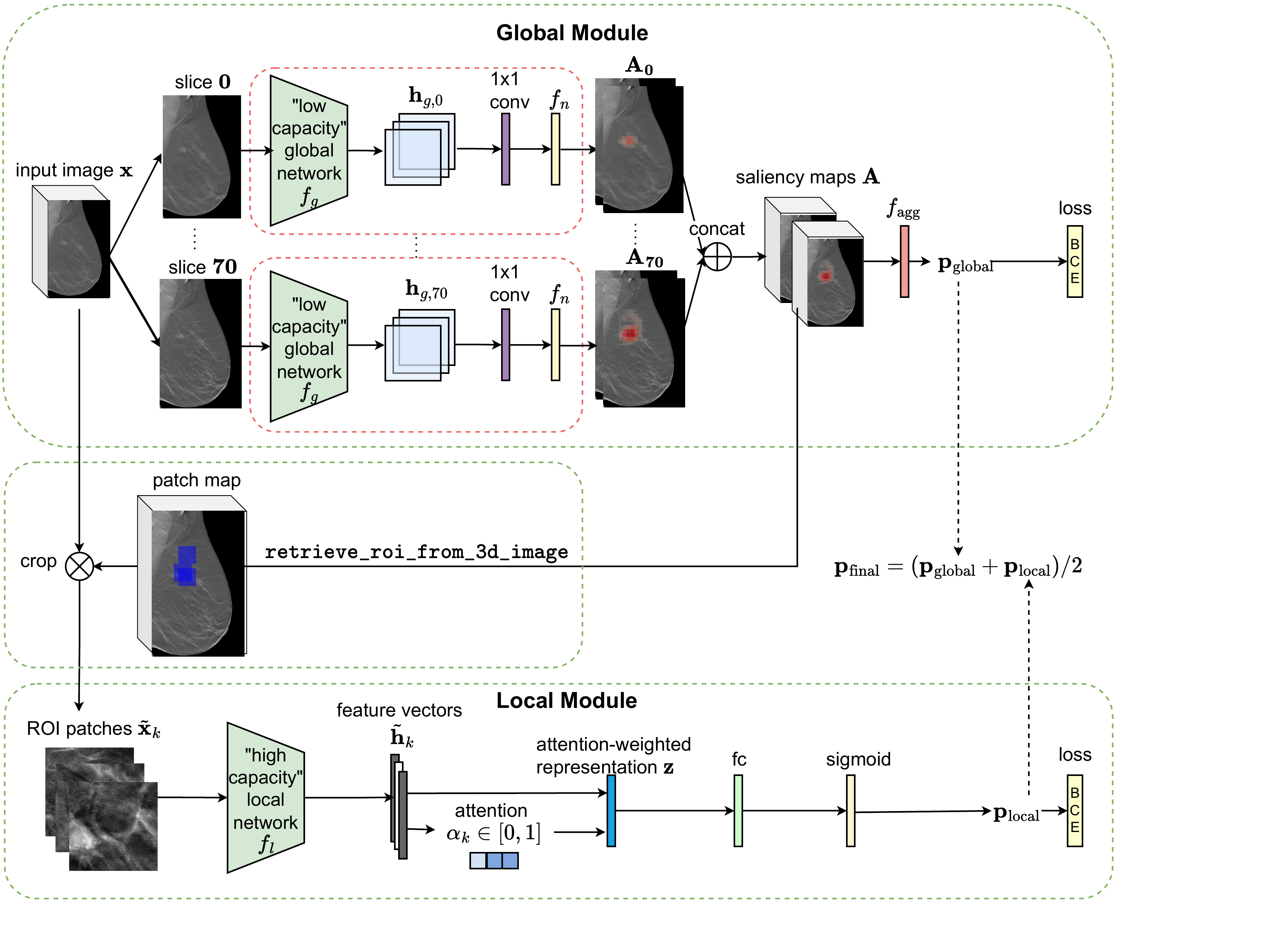}
  \vspace{-7.5mm}
  \caption{Architecture of 3D-GMIC. 3D-GMIC first applies the same low-capacity global network to all slices of a DBT image to generate the saliency map. 
  The $\texttt{retrieve\_roi\_from\_3d\_image}$ algorithm crops ROI patches from the regions corresponding to the highest values in the saliency map.
  The model then applies the high-capacity local network to the cropped patches to focus the computation in the sampled regions.
  Finally, the model combines the predictions from the global and local modules to create the overall prediction for the input image.
  }
  \label{fig:overall_plot}
\vspace{-3mm}
\end{figure*}

The proposed model, 3D-GMIC (Fig.~\ref{fig:overall_plot}), extends GMIC~\cite{shen2021interpretable} to 3D data.
The low-capacity global network $\mathbf{f_g}$ is a convolutional neural network which processes 2D input images. 
As in the original GMIC, the global network $\mathbf{f_g}$ is parameterized as ResNet-22~\cite{wu2019deep} which is narrower and has larger strides compared to the canonical ResNet architectures~\cite{he2015deepresidual}.
3D-GMIC applies the global network separately to each slice of a 3D image in parallel. 
For each slice $\mathbf{x_d}$ in the input 3D image $\mathbf{x}$ with $D$ slices, the global network first extracts the hidden representation $\mathbf{h_{g,d}} \in \mathbb{R}^{h,w,c}$ where h,w,c are the hidden dimensions.
The hidden representation $\mathbf{h_{g,d}}$ is turned into saliency maps $\mathbf{A_d}$ using a semantic segmentation layer, which is a combination of a 1x1 convolution layer with a nonlinear function $\mathbf{f_n}$.
Concretely,
\begin{equation}
    \mathbf{A_d} =  f_n(\text{conv}_{1\times1}(\mathbf{h}_{g,d})).
\end{equation}
The saliency maps for all slices of $\mathbf{x}$ are then aggregated to form the 3D saliency map $\mathbf{A} \in \mathbb{R}^{h,w,D,2}$.
These saliency maps identify the most important regions and slices of the input image $\mathbf{x}$ for the benign and malignant categories.

Then, for each class $c \in \{b, m\}$, we use an aggregation function $f_\text{agg}(\mathbf{A}^c): \mathbb{R}^{h,w,D} \mapsto [0,1]$ to transform the saliency map $\mathbf{A^c}$ into image-level class prediction $\mathbf{p}_\text{global}^c$:
\begin{equation}
    \mathbf{p}_\text{global}^c = f_\text{agg}(\mathbf{A}^c).
\end{equation}
We define the aggregation function as
\begin{equation}
    f_{\text{agg}}(\mathbf{A}^c) = \frac{1}{|H^+|}\sum_{(i,j,d) \in H^+} \mathbf{A}^c_{i,j,d},
\end{equation}
where $H^+$ denotes the set containing locations of top $t\%$ values in $\mathbf{A}^c$ as described in section~\ref{sec:pooling}. 

\subsubsection{The local module} \label{sec:local}
In \textit{the local module} of 3D-GMIC, we select the most important regions of $\mathbf{x}$ according to the saliency maps $\mathbf{A}$. 
Concretely, we greedily select $K$ square patches which corresponds to the high values in the saliency maps as in the $\texttt{retrieve\_roi}$ algorithm from Shen et al.~\cite{shen2021interpretable}.
We adapt and modify this algorithm for 3D images to prohibit selecting a patch if one of the previously selected patches were at the same xy-location and from $\pm \zeta$ neighboring slices.
This prevents cropping multiple patches with duplicate information from nearby slices.
We name this algorithm $\texttt{retrieve\_roi\_from\_3d\_image}$ (Algorithm~\ref{alg:roi}).
The width and height of the patches are fixed to 256 in all experiments.
This algorithm identifies $K$ image patches $\{\tilde{\mathbf{x}}_k\}$ to maximize the criterion in line 7 at each selection as follows:
    $\{\tilde{\mathbf{x}}_k\} = \texttt{retrieve\_roi\_from\_3d\_image}(\mathbf{A})$.

We can then apply the high-capacity local network $\mathbf{f_l}$ to utilize fine-grained details from the $K$ selected image patches $\{\tilde{\mathbf{x}}_k\}$ by computing
    $\tilde{\mathbf{h}}_k = f_l(\tilde{\mathbf{x}}_k)$. 
The extracted feature vectors $\{\tilde{\mathbf{h}}_k\}$ are then aggregated using a gated attention mechanism~\cite{ilse2018attention}.
Attention scores, $\alpha_k \in [0,1]$, indicating the relevance of each patch are calculated as follows:
\begin{equation}
    \alpha_k = \frac{\text{exp}\{\mathbf{w}^\intercal (\text{tanh}(\mathbf{V}\mathbf{\tilde{h}}_k^{\intercal}) \odot \text{sigm}(\mathbf{U}\mathbf{\tilde{h}}_k^{\intercal}) )\}}{\sum^K_{j=1}\text{exp}\{\mathbf{w}^\intercal (\text{tanh}(\mathbf{V}\mathbf{\tilde{h}}_j^{\intercal}) \odot \text{sigm}(\mathbf{U}\mathbf{\tilde{h}}_j^{\intercal}) )\}},
\end{equation}
where $\odot$ denotes an element-wise multiplication, $\mathbf{w} \in \mathbb{R}^{L \times 1}$, $\mathbf{V} \in \mathbb{R}^{L \times S}$ and $\mathbf{U} \in \mathbb{R}^{L \times S}$ are learnable parameters. 
In all experiments, we set $L = 128$ and $S = 512$.
This process yields an attention-weighted representation
\begin{equation}
    \mathbf{z} = f_a(\{\tilde{\mathbf{h}}_k\}) = \sum_{k=1}^{K} \alpha_k \tilde{\mathbf{h}}_k.
\end{equation}

We then apply a fully connected layer with sigmoid nonlinearity to $\mathbf{z}$ to generate the prediction 
    $\mathbf{p}_{\text{local}} = \text{sigm}(\mathbf{w_{\text{local}}}^T \mathbf{z})$,
where $ \mathbf{w}_{\text{local}} \in \mathbb{R}^{S \times 2}$ are learnable parameters.

\subsubsection{3D-GMIC output} 

We average the predictions from the global and local modules of 3D-GMIC to produce the final class predictions as  $\mathbf{p}_{\text{final}}^c = (\mathbf{p}_{\text{global}}^c + \mathbf{p}_{\text{local}}^c) / 2$.

\subsubsection{The loss function used in the training of 3D-GMIC} \label{sec:learning}

3D-GMIC is trained end-to-end by minimizing binary cross-entropy (BCE) losses for the predictions from the two stages $\mathbf{p}_{\text{global}}^c$ and $\mathbf{p}_{\text{local}}^c$.
In addition, we encourage sparsity on the saliency maps $\mathbf{A}$ by imposing $L1$ regularization $L_\text{reg}(\mathbf{A})$:
\begin{equation}
L_\text{reg}(\mathbf{A}^c) = \sum_{(i,j,d)} |\mathbf{A}^c_{i,j,d}|.
\end{equation}
This enables successful semantic segmentation of important regions in the saliency map. In summary, the training loss has the following form:
\begin{equation}
    \begin{split}
        L(\mathbf{y}, \mathbf{p}) = \sum_{c \in \{b,m\}} ( \text{BCE}(\mathbf{y}^c, \mathbf{p}_{\text{local}}^c) + \text{BCE}(\mathbf{y}^c, \mathbf{p}_{\text{global}}^c) \\ + \beta L_\text{reg}(\mathbf{A}^c)),
    \end{split}
\end{equation}
where $\beta$ is a hyperparameter. 
We calculate the two BCE losses separately and sum them up rather than calculating one BCE loss with the arithmetic average between the two predictions ${\mathbf{y}}_{\text{global}}^c$ and ${\mathbf{y}}_{\text{local}}^c$.
This is because the latter leads to underutilization of one of the modules~\cite{wu2020improving}.
In other words, optimizing the latter BCE loss leads to one of the modules learning to predict the cancer accurately and the other module just predicting the same probability for all images.

\subsection{Handling small datasets with large images}

DBT is a relatively recent imaging modality and therefore researchers might have a limited number of DBT exams available.
In addition, DBT is an imaging modality consisting of hundreds of millions of pixels and training AI models on such large images is technically challenging.
In this section, we describe the components which enable processing DBT images without compromising performance.

\subsubsection{Group Normalization}
For the global network, we replace batch normalization~\cite{ioffe2015batch} with group normalization~\cite{wu2018group} with 8 groups to retain performance with small batch size and avoid treating each slice of image as different examples in batch normalization. 
Since group normalization is mostly robust to the choice of the number of groups, we choose 8 as it showed good performance in the original paper and all channel sizes are divisible by 8. %

\subsubsection{Patch sampling in 3D} \label{sec:sampling3d}
In DBT, a lesion appears across multiple slices with varying focus. 
Cropping patches from all of these slices is unnecessary as they are redundant.
Therefore, the $\texttt{retrieve\_roi\_from\_3d\_image}$ algorithm (Algorithm~\ref{alg:roi}) is designed to avoid cropping redundant patches.
Specifically, when cropping each patch, we remove saliency from nearby slices at the same xy-location as the best candidate patch to prevent cropping redundant patches in the following steps.
In addition, during training, we uniformly sample one of the patches from nearby slices ($\pm\zeta$ slices) at the same xy-location as a part of input data augmentation.

\renewcommand{\algorithmicrequire}{\textbf{Input:}}
\renewcommand{\algorithmicensure}{\textbf{Output:}}
\begin{algorithm}[t]
    \caption{\texttt{retrieve\_roi\_from\_3d\_image}}
    \label{alg:roi}
    \begin{algorithmic}[1]
        \REQUIRE  $\mathbf{x} \in \mathbb{R}^{H,W,D}$, $\mathbf{A} \in \mathbb{R}^{h,w,D,2}$, $K$
        \ENSURE $ O = \{ \tilde{\mathbf{x}}_k |  \tilde{\mathbf{x}}_k \in \mathbb{R}^{h_c,w_c,D} \}$
        \STATE{$O = \emptyset$}
        \FOR{each class $c \in \{\text{benign}, \text{malignant}\}$}
            \STATE{$\mathbf{\tilde{A}}^c = \text{min-max-normalization}(\mathbf{A}^c)$}
         \ENDFOR\\
         \STATE{$ \mathbf{A}^{*} = \sum_{c} \tilde{\mathbf{A}}^c$}
         \STATE{$l$ denotes an arbitrary $h_c \frac{h}{H} \times w_c \frac{w}{W}$ rectangular patch on $\mathbf{A}^{*}$}
         \STATE $\text{criterion}(l, \mathbf{A}^{*}) = \sum_{(i,j) \in l} \mathbf{A}^{*}[i,j]$
        \FOR{each $1,2,...,K$}
            \STATE{$l^* = \argmax_{l} \text{criterion}(l, \mathbf{A}^{*})$}
            \STATE{$L = $ position of $l^*$ in $\mathbf{x}$}
            \IF{(D>1) and (model.training)}
                \STATE{L = uniformly sample a patch from the $\pm \zeta$ neighboring slices of L}
            \ENDIF\\
            \STATE{$O = O \cup \{L\}$}
            \STATE{$Z$ = area including $l^* \pm \zeta$  neighboring slices}
            \STATE{ $\forall (i,j) \in Z$, set $\mathbf{A}^*[i,j]=0$}
        \ENDFOR
        \STATE \textbf{return} $O$ 
    \end{algorithmic}
\end{algorithm}

\subsubsection{Saliency aggregation independent of image slices} \label{sec:pooling}
For the benign and malignant classes, the aggregation function $f_{agg}$ in \textit{the global module} of standard GMIC pools the top $t\%$ of the values from the saliency maps $\mathbf{A^b}$ and $\mathbf{A^m}$.
It can be viewed as a balance between global average pooling and global max pooling. 
In the standard GMIC, this top t\% pooling leads to superior performances in comparison to either of the extremes.

The size of the saliency maps $\mathbf{A^b}$ and $\mathbf{A^m}$ for a DBT image is $h \times w \times D$ where $h$ is the height, $w$ is the width, and $D$ is the number of slices in the input image (therefore, also in the saliency map).
If the aggregation function $f_{agg}$ pools $t\%$ of the pixels with respect to the entire $\mathbf{A^b}$ and $\mathbf{A^m}$, then the number of pooled pixels in each saliency map is $t/100 \times h \times w \times D$.
Since this formula is dependent on $D$, the aggregation function $f_{agg}$ pools different numbers of pixels from the saliency maps from images with different numbers of slices.
However, the sizes of lesions in DBT images do not depend on the number of slices.\footnote{The thickness of a compressed breast completely determines the number of slices in the corresponding DBT image. Specifically, for the NYU Combo v1 dataset, the number of slices of a DBT image is equal to the thickness of the corresponding compressed breast in millimeters + 6. For example, if a compressed breast is 55mm thick, then its DBT image has 61 slices.} 
Since the sizes of the lesions are independent of the sizes of the breasts, the formula's dependence on $D$ can cause contradictory training signals and outputs.

For example, suppose there exist two DBT images that contain an identical malignant lesion but consist of different numbers of slices.
Since the lesions show the same characteristics such as shape and size, it will highlight the same number of pixels in the saliency maps corresponding to these two images.
Nonetheless, the aggregation function $f_{agg}$ will pool different number of pixels from these saliency maps because they have different number of slices.
This can lead to the predictions of malignancy of \textit{the global module} $\mathbf{p}_{\text{global}}^m$ to greatly differ between these two images even though the lesions are identical.
This inconsistency could make the learning more difficult than necessary and degrade the performance.

Ideally, $\mathbf{p}_{\text{global}}^m$ should not depend on the size of the breast and only depend on the detected lesions.
To do so, 3D-GMIC applies $f_{agg}$ to the saliency maps $\mathbf{A}$ to pool the values in a way that does not depend on the number of slices $D$ in the input image $\mathbf{x}$.
Specifically, we define the pooling percentage $t\%$ with respect to a single slice of a 3D image.
For example, from saliency maps of size $h \times w \times 50$ and size $h \times w \times 80$, setting $t=200\%$ will pool $2 \times h \times w$ values from both saliency maps which accounts for 4\% and 2.5\% of the entire 3D saliency map, respectively.
This is still a small subset of the entire 3D saliency maps, which allows 3D-GMIC to localize the important regions.
At the same time, the number of pooled values from the saliency maps no longer changes across images with different number of slices.

\subsubsection{The initialization of the semantic segmentation layer}~\label{sec:seg_layer}

\begin{figure}[ht]
\centering
\includegraphics[scale=0.5, trim={1.25cm 0 1cm 0}]{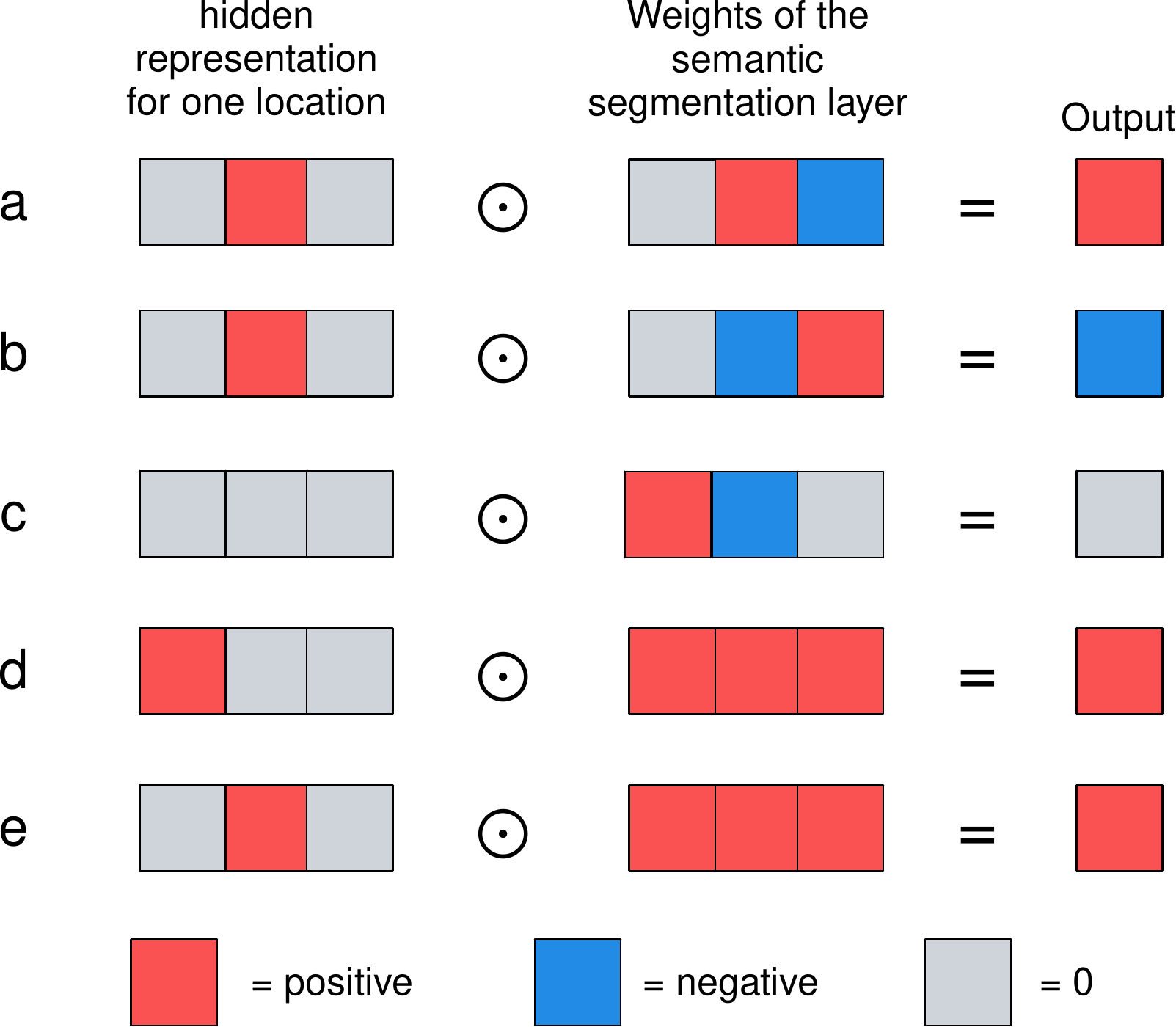}

\caption{
The problem with randomly initialized weights of the 1x1 convolution filter in the semantic segmentation layer.
  Randomly initialized weights \textbf{(a-c)} can map high values of the hidden representation $h_g$ to negative outputs \textbf{(b)}.
  This leads potentially useful features to have lower values in the saliency maps than those from the background.
  In comparison, weights initialized with a constant $\omega$ \textbf{(d-e)} always map high hidden representation values to positive output.
  This ensures that the values in the saliency maps from interesting regions will be higher than those from the background, which is a more reasonable starting point for model training and leads to consistent success of semantic segmentation.
}
\label{fig:randinit}
\vspace{-3mm}
\end{figure}

In the \textit{the global module} of standard GMIC, the weights of the semantic segmentation layer are randomly initialized. 
However, we find that this often leads to the failure of weakly-supervised semantic segmentation in low-data regimes such as in this work\footnote{Even though the training set consists of 78,702 exams from 68,412 patients, it contains only 518 breasts with malignant findings, therefore we say this is low-data regime.}, as shown in Fig.~\ref{fig:randinit}. 

The problem with randomly initialized weights in the convolutional layer occurs when performing transfer learning from a pretrained ReLU network with top t\% pooling, as shown in Fig.~\ref{fig:randinit}.a\textasciitilde{}c. 
The pretrained CNN with ReLU nonlinearity tends to output high positive logit values in some channels for locations where small, important features are. 
Ideally, this value should be multiplied with some positive weight in the semantic segmentation layer such that we can preserve this information as shown in Fig.~\ref{fig:randinit}.a.
With random initialization, however, negative weights often get assigned to these channels which detect important features as shown in Fig.~\ref{fig:randinit}.b.
The outputs from such important features will then be lower than the output of the unimportant regions shown in Fig.~\ref{fig:randinit}.c.

\begin{figure}[ht]
\centering
\includegraphics[width=0.22\textwidth, trim={0 0 0 0}]{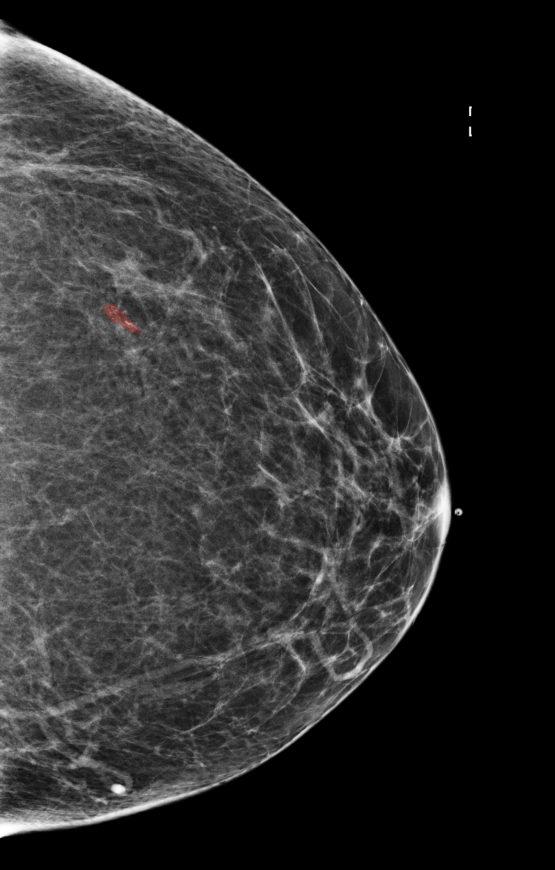}
\includegraphics[width=0.22\textwidth, trim={0 0 0 0}]{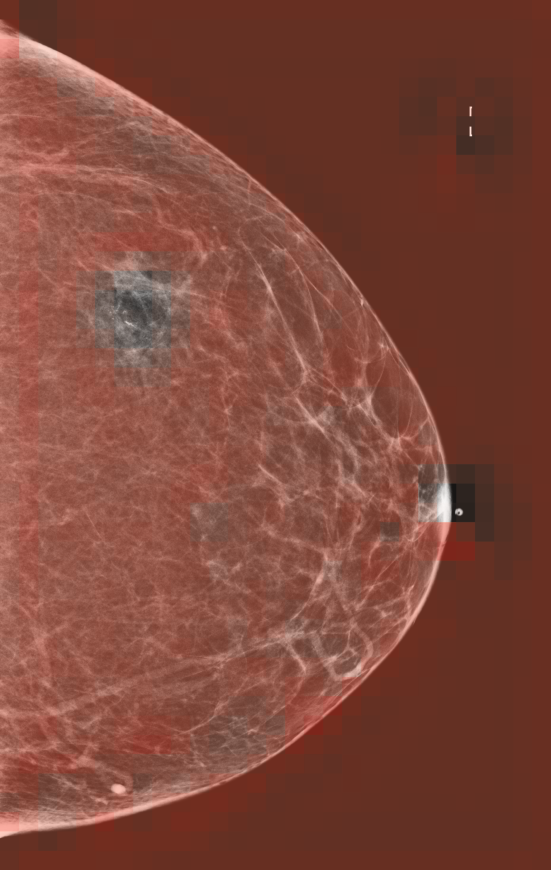}

\caption{
Failure of weakly-supervised semantic segmentation.
As explained in section~\ref{sec:seg_layer}, the areas in the saliency map \textbf{(right)} corresponding to the biopsied lesion \textbf{(left)} have lower values than the unimportant regions and the background after BI-RADS pretraining.
In this case, it is difficult for model to learn to highlight lesions correctly with top-t\% pooling.
}
\label{fig:failure_case}
\vspace{-3mm}
\end{figure}

In such a case, the important regions will not be pooled to $\mathbf{p}_{\text{global}}^c$ as shown in Fig.~\ref{fig:failure_case}.
Instead, $f_{agg}$ will pool from the background and normal regions without any pathology since the values in the saliency maps corresponding to these regions will be higher than those where malignant or benign lesions are.
As a result, the saliency pooled from normal regions from malignant breasts will be encouraged to output 1, teaching the model to highlight the wrong portions of the images.
This leads to the failure of weakly-supervised semantic segmentation.

To address this issue, we initialize the $\text{conv}_{1\times1}$ layer in the semantic segmentation layer with constant, positive weights $\omega$ (Fig.~\ref{fig:randinit}.d\textasciitilde{}e). 
No matter which channel of the pretrained model reacts to important regions in the image, it will be correctly captured as high value in the saliency maps $\mathbf{A}$ since the model weighs all channels equally at the beginning.
In this configuration, $f_\text{agg}$ will pool from important regions in positive exams and encourage the pooled values to be closer to 1. 

\subsubsection{The choice of nonlinearity $f_n$}

When using constant initialization of the semantic segmentation layer, sigmoid function is no longer an ideal choice for nonlinearity $f_n$.
For the sigmoid function to suppress the probability predictions corresponding to background regions, the 1x1 convolution layer in the semantic segmentation layer must output highly negative values for uninteresting regions.
When using the constant weight initialization and sigmoid nonlinearity, the low logits from the background regions are mapped close to 0.5 probability in the saliency map, and the high logits from the suspicious regions are mapped to probability values that are somewhat higher than 0.5 at the beginning of training.
To decrease the 0.5 probability predictions from the background regions, the backbone network $f_g$ must learn to output non-zero logits for some channels from all such background regions and then the 1x1 convolution layer must associate negative weights for such logits.
However, if the model had not already learned to do so during the pretraining phase, it could be hard to learn when fine-tuning with the downstream task.
To learn this behavior during the fine-tuning, the pretrained backbone network must change significantly, which could risk catastrophic forgetting.

\begin{figure}[ht]
\centering

\begin{tabular}{cc@{\hspace{\tabcolsep}}c@{\hspace{\tabcolsep}}c}
nonlinearity & $\omega$=0.001 & $\omega$=0.01 & $\omega$=0.05\\ \hline
\includegraphics[scale=0.25]{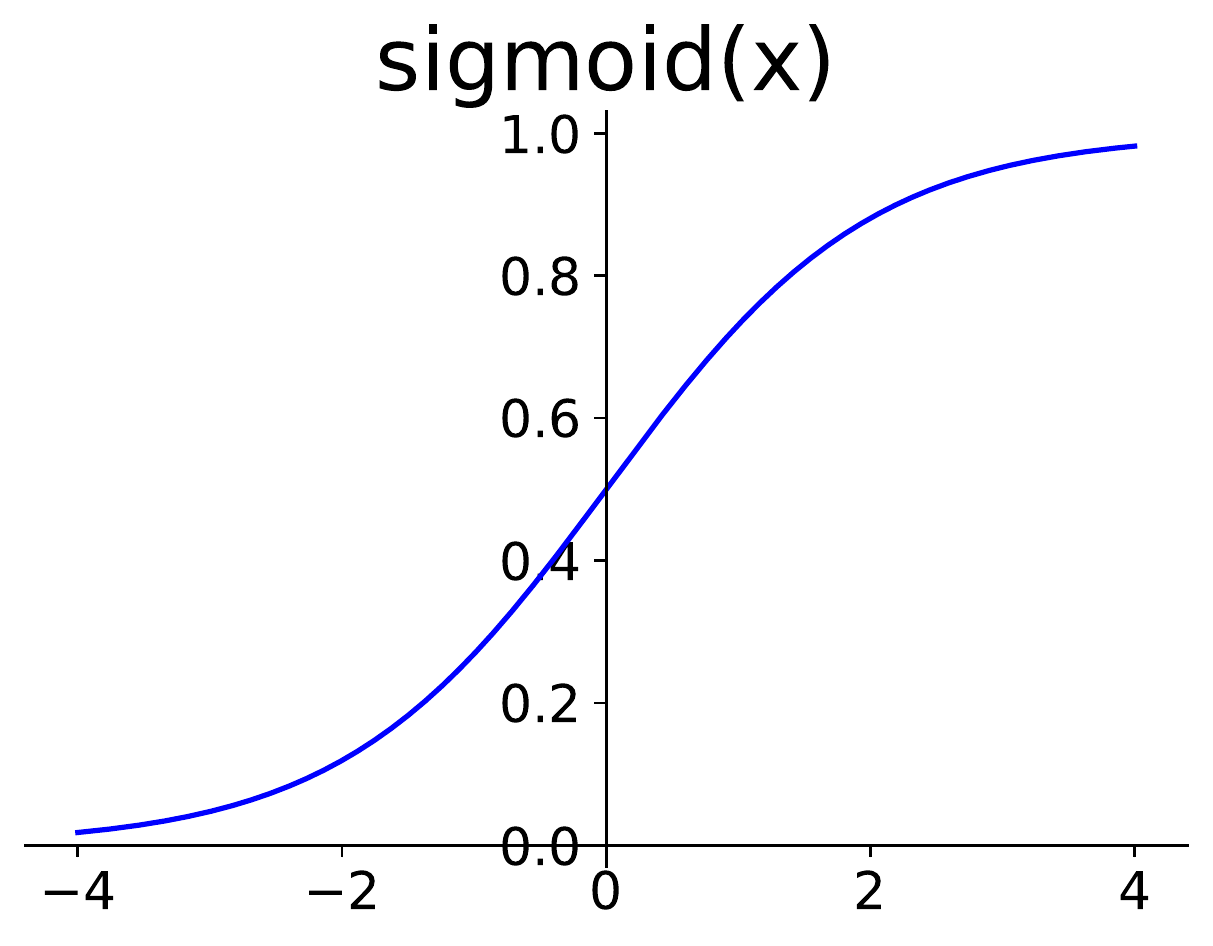} &
\includegraphics[scale=0.12]{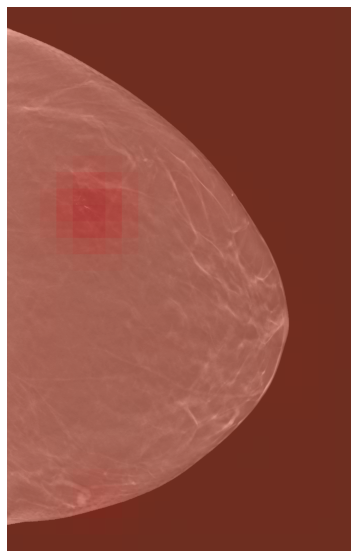}&
\includegraphics[scale=0.12]{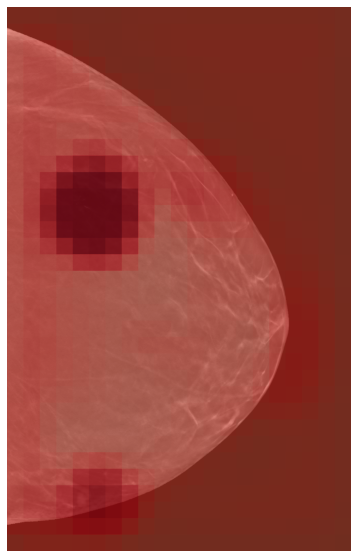}&
\includegraphics[scale=0.12]{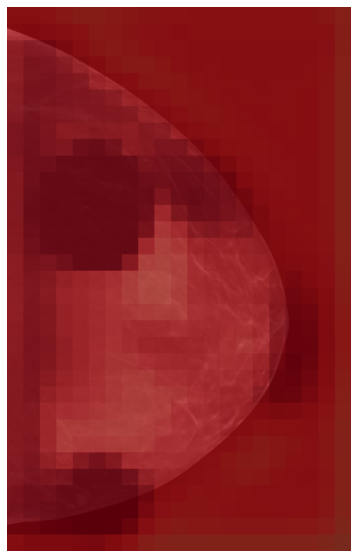}\\
\includegraphics[scale=0.25]{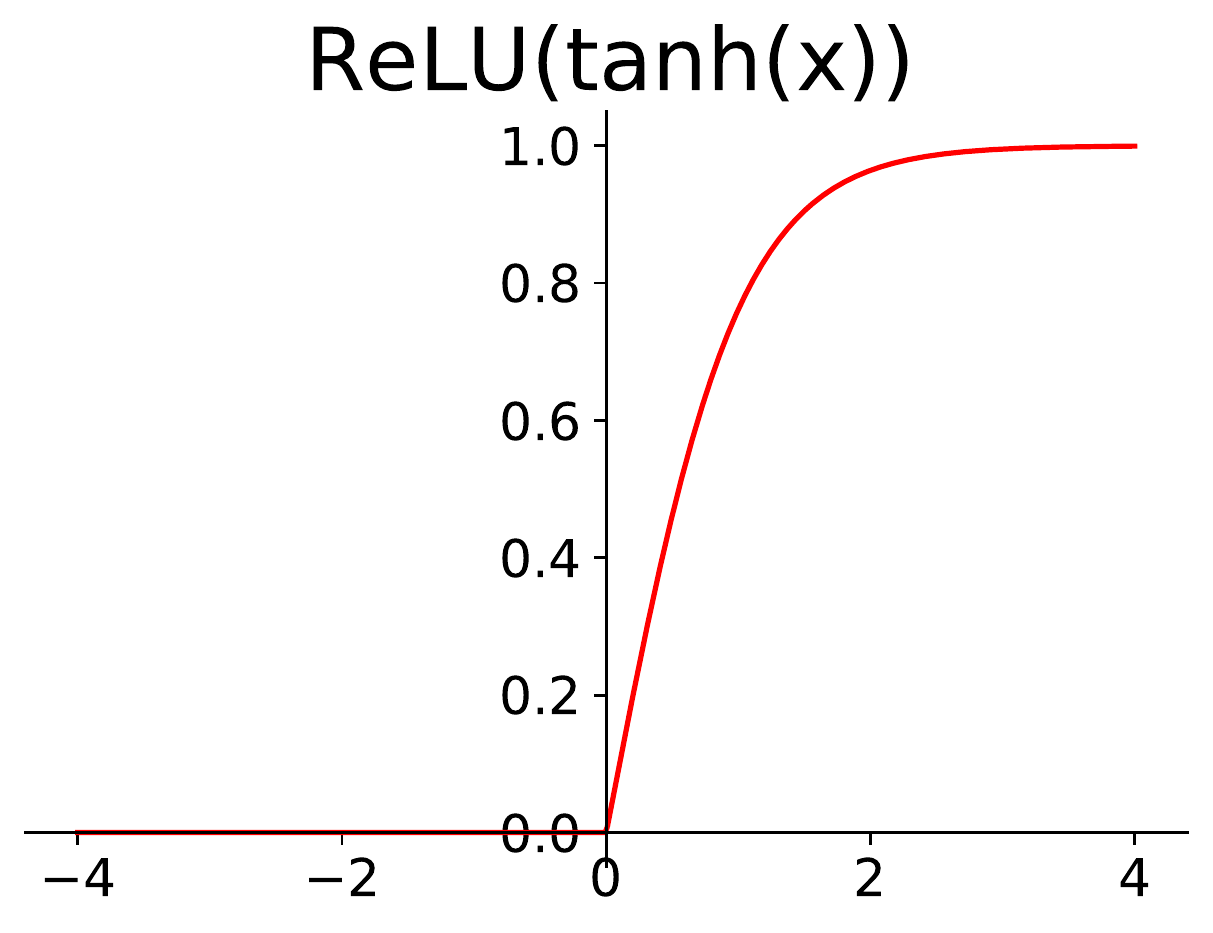} &
\includegraphics[scale=0.12]{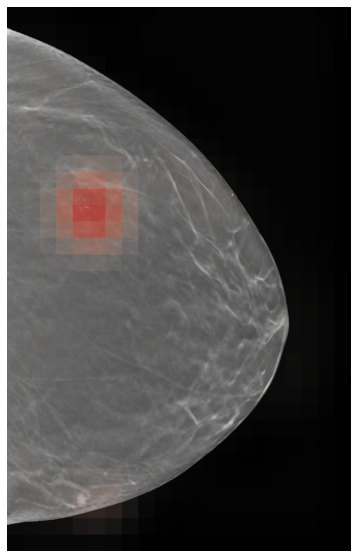}&
\includegraphics[scale=0.12]{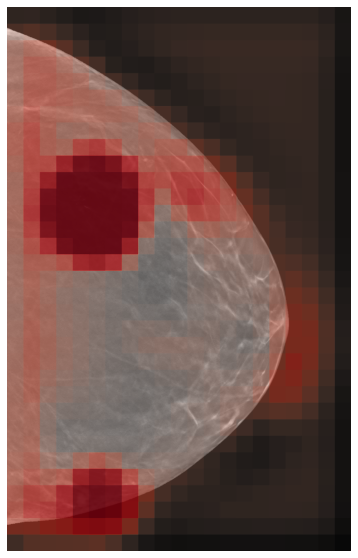}&
\includegraphics[scale=0.12]{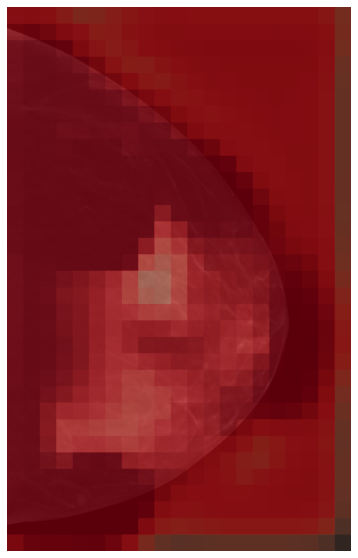}
\end{tabular}
\caption{The effect of different nonlinearity functions and the weight initialization constant $\omega$ on a global module pretrained with BI-RADS labels.
The visualizations of saliency maps are made with different initialization constants $\omega$ before any further training on the pathology labels.
Sigmoid \textbf{(top)} leads to initially predicting 0.5 everywhere in the background, which requires the model to learn to suppress saliency in the background in the downstream training.
On the other hand, ReLU(tanh(x)) \textbf{(bottom)} maps all the unimportant near-zero hidden representations to near-zero in the saliency maps from the beginning, which is an easier starting point that ensures the success of semantic segmentation.
The effect of three different values of $\omega$ for each nonlinearity is shown in the right panel.
With the pretrained global module, we observe that $\omega$ between 0.001 and 0.1 results in saliency maps that sparsely highlight the important regions.
} 
\label{fig:topk}
\vspace{-3mm}
\end{figure}

To mitigate this issue, an ideal nonlinearity $f_n$ should naturally map the near-zero outputs of the 1x1 convolution layer into zero probabilities in the saliency maps.
One might think that shifting the sigmoid function horizontally could achieve this.
However, the input interval corresponding to the highest rate of change in nonlinearity also shifts away from 0, which hinders learning.
The ideal nonlinearity for constant weight initialization would map near-zero values to 0 and would have the highest rate of change near zero.

To satisfy these criteria, we propose ReLU(tanh(x)) for $f_n$ as shown in Fig.~\ref{fig:topk}.
Like sigmoid, ReLU(tanh(x)) also maps all input values to the $[0, 1]$ interval.
Unlike sigmoid, ReLU(tanh(x)) turns the near-zero outputs of the 1x1 convolution layer into zero values in the saliency map.
At the same time, the highest rate of change occurs with the input values around zero.
This simplifies the learning and no longer requires a change of behavior in the backbone network $f_g$.
We observe that ReLU(tanh(x)), combined with the constant initialization of the semantic segmentation layer, leads to consistent success of weakly-supervised semantic segmentation in the saliency maps $\mathbf{A}$ with the NYU Combo v1 dataset which is smaller than the dataset used in Shen et al.~\cite{shen2021interpretable}.

\subsection{Fusion module}

The standard GMIC includes a fusion module to combine the hidden representations of the global and local modules to output the final prediction.
In the preliminary experiments, however, we found that the fusion module does not provide benefit with DBT and ReLU(tanh(x)) nonlinearity, and thus we exclude it. 
We hypothesize that this is because more information is lost in max pooling from 3D representations than max pooling from 2D, which might prevent the learning.

\subsection{Training procedure}

\begin{figure}[ht]
\centering
\vspace{3mm}
\includegraphics[scale=0.7]{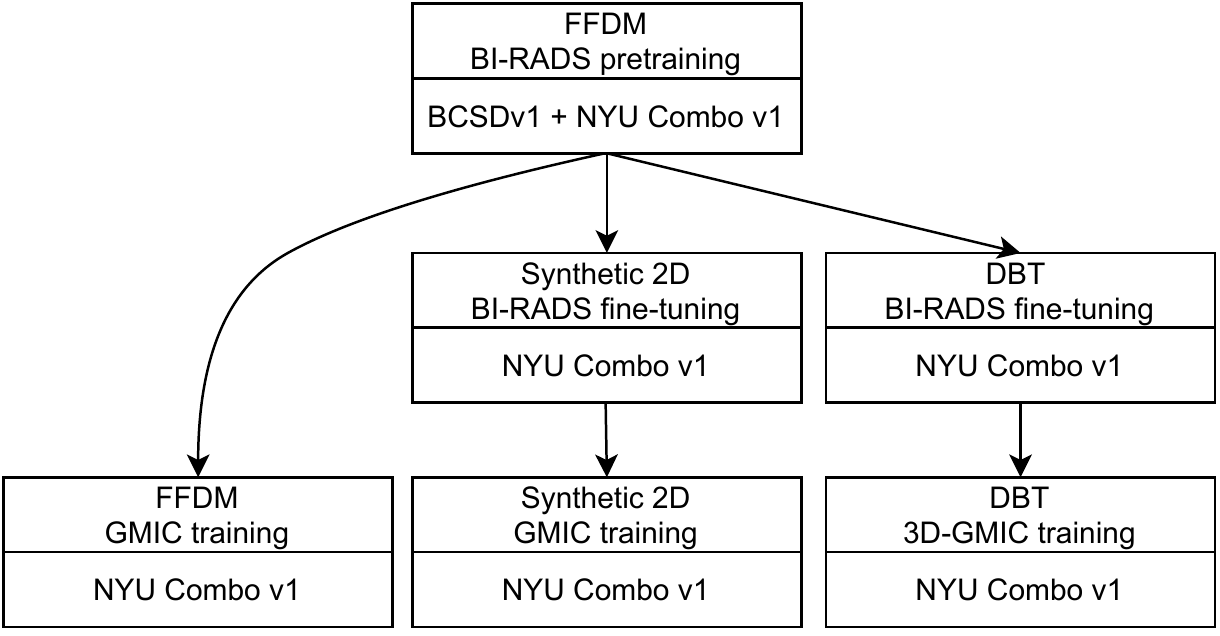}

\caption{
The training pipeline for the three imaging modalities.
}
\vspace{-3mm}
\label{fig:pipeline}
\end{figure}

To train the model architectures on different modalities we take the following steps, as shown in Fig.~\ref{fig:pipeline}:
\begin{enumerate}
    \item Pretrain the global network on BI-RADS labels with the BCSDv1+NYU Combo v1 combined dataset using FFDM modality only (described in section~\ref{sec:birads}).
    \item Transfer the model from (1) to synthetic 2D or DBT modalities in the NYU Combo v1 dataset, and further train on the BI-RADS classification task to fine-tune the pretrained weights to each modality. 
    \item Train the FFDM model by transferring the pretrained model weights from (1).
    \item Train the synthetic 2D and DBT models by transferring the pretrained model weights from (2).
\end{enumerate}

We compare performances from models created in steps 3 and 4 in Table~\ref{tab:auc_table}, Table~\ref{tab:specificity_table}, and Table~\ref{tab:mcc_table}.

\subsubsection{BI-RADS pretraining} \label{sec:birads}
For the global network of our models, we apply transfer learning from networks pretrained with BI-RADS labels as described by Wu et al.~\cite{wu2019deep} and Geras et al.~\cite{geras2017high}.
We modify the BI-RADS pretraining procedure so that instead of predicting BI-RADS labels utilizing all of the four views in an exam at once, we predict BI-RADS labels for each image separately.
We first pretrain a model parameterized with ResNet-22 architecture~\cite{wu2019deep} using the FFDM images, and transfer to further BI-RADS pretraining on other modalities and training GMIC models as described above.
For the pretraining phase for the models with DBT images, we use the same ResNet-22 model but apply it to all slices in each image in parallel.
And then, we aggregate the average-pooled representation from all slices using a gated attention mechanism~\cite{ilse2018attention} before applying logistic regression.



\subsubsection{Training 2D GMIC}
 
We use GMIC for training models with FFDM and synthetic 2D data.
For a fair comparison, we apply the same modifications made to 3D-GMIC when we can.
Specifically, we apply the same ReLU(tanh(x)) nonlinearity for the generation of the saliency map, replace the fusion module with the average prediction between global and local modules, and replace batch normalization with group normalization in \textit{the global module}.
While these changes do not necessarily improve the best possible performances with 2D-GMIC, our initial experiments indicated that applying them makes learning more stable when learning with fewer data.

\subsubsection{Image Augmentation}
FFDM images have dimensions (4096, 3328) or (3328, 2560).
DBT and synthetic 2D image sizes are either (2457, 1996) or (2457, 1890). 
The corresponding FFDM, DBT, and synthetic 2D images have the same field of view.
We crop each image to a predefined input size using the procedure described by Wu et al.~\cite{wu2019deep}. 
FFDM images are cropped to the size of 2866 $\times$ 1814 pixels, whereas synthetic 2D images and DBT images are cropped to the size of 2116 $\times$ 1339 pixels to capture the equivalent field of view. 
These window sizes were selected as they yield an equivalent field of view for all modalities, considering that the original image resolutions differ between modalities.
We also apply random shifting and resizing during training and testing phases for a maximum of 100 pixels in any direction.
At test time, we apply 10 random augmentations to get 10 predictions for each image and average them to make a final prediction.

\subsubsection{Mixed-precision training}
To decrease GPU RAM usage, we use the Apex library\footnote{\url{https://github.com/NVIDIA/apex}} for mixed-precision training~\cite{micikevicius2017mixed}.

\subsubsection{Distributed training}
To maximize the training speed and increase batch size, we parallelize the training over 4 Nvidia v100 GPUs when training models on the 3D modality. 
The effective batch size per update is 4 images.
For \textit{the local module}, we use a synchronized batch normalization layer to share batch statistics between GPUs.

\subsubsection{Hyperparameter tuning}

We optimize network hyperparameters with random search~\cite{bergstra2012random}. 
For all models, we randomly sample the learning rate $\eta$ from log-uniform distribution $\log_{10}(\eta) \sim \mathcal{U}(-5.5, -4.5)$  and the initialization constant $\omega$ from log-uniform distribution $\log_{10}(\omega) \sim \mathcal{U}(-3, -2)$.

For 2D GMIC, we randomly sample the pooling percentage $t$ with respect to the entire image from uniform distribution $t \sim \mathcal{U}(1\%, 25\%)$, the number of patches $K \in \{4, 6, 8\}$ with equal probabilities, and the regularization weight $\beta$ from log-uniform distribution $\log_{10}(\beta) \sim \mathcal{U}(-5.5, -3.5)$.

For 3D-GMIC, we choose a different range of some hyperparameters compared to 2D to mitigate the differences between the imaging modalities. 
For example, the lesions in the DBT contain 10.97 times more pixels on average compared to the lesions in synthetic 2D images, regardless of the number of slices in DBT images. 
As this suggests that more salient pixels will be highlighted in the saliency maps for DBT images than the corresponding synthetic 2D images, we increase the minimum and maximum value of the pooling percentage $t$ 10.97 times to capture a comparable amount of highlighted area in the saliency maps $A$.
In addition, we decrease the minimum and maximum value of regularization weight $\beta$ 10.97 times so that the regularization term has a comparable magnitude to the remaining terms in the final loss function used to train 3D-GMIC.


Finally, we adjust the number of sampled patches $K$ for 3D-GMIC to account for the depth in DBT data.
Even though there are about 70 slices on average, we do not increase $K$ by 70 times compared to GMIC on 2D images.
This is because our method is more efficient than applying GMIC on all slices in parallel.
Since our $\texttt{retrieve\_roi\_from\_3d\_image}$ algorithm avoids cropping duplicate information from nearby patches, 3D-GMIC can focus its computation on much smaller proportion of the input image compared to GMIC.
Nonetheless, as a precaution, we increase the number of patches $K$ by a conservative factor of 2 compared to GMIC on 2D images as some patches might still be located within the same xy coordinates but at different slices.
In such a case, the total number of xy-coordinates from which patches are sampled in 3D-GMIC could be decreased compared to GMIC on 2D mammography, and we want to mitigate this.
In Table~\ref{tb:ablation_numpatches}, we show that this increase was not strictly necessary, as both GMIC and 3D-GMIC utilizing the same number of patches achieve comparable performances.

As a result, we randomly sample the pooling percentage $t$ from uniform distribution $t \sim \mathcal{U}(10.97\%, 274.25\%)$ with respect to one slice of an image\footnote{Even though the pooling percentage can be larger than 100\% of a slice of a saliency map, it is still a small subset of the entire 3D saliency map.} from uniform distribution, number of patches in image $K \in \{8, 12, 16\}$ with equal probabilities, and the regularization weight $\beta$ from log-uniform distribution $\log_{10}(\omega) \sim \mathcal{U}(-6.54, -4.54)$.

For all models described in this paper, we optimize their hyperparameters with 40 random search trials. 
All models were trained for a maximum of 40 epochs but the ones which do not show any improvement in validation performances for 15 consecutive epochs are early stopped. 
We selected 5 models with the highest AUC scores in identifying images with malignant findings on the validation set.

Training 3D-GMIC for 40 epochs takes about three days when utilizing four Tesla V100 GPUs with 32GB GPU memory in parallel.
Training GMIC with FFDM and synthetic 2D data for 40 epochs on a Tesla V100 GPU with 16GB GPU memory takes about 18 hours and about 12 hours, respectively.


\section{Evaluation Metrics} \label{section:eval}

\subsection{Computational efficiency}
To compare 3D-GMIC and other architectures, we measure the maximum GPU memory usage during training and the number of computations in Giga Multiply-ACcumulate operations (GMACs) by benchmarking with a single DBT image with 96 slices.
Since it is not possible to run off-the-shelf models with all of 96 slices, we estimate their metrics using linear extrapolation with DBT images with 4, 8, 16 slices. 

\subsection{Classification}
To evaluate the classification performance, we calculate area under the receiver operating characteristic curve (AUC) on two levels of granularity:
\begin{enumerate}
    \item Image-wise AUC, where the performance of the model is evaluated using the predictions for each image.
    \item Breast-wise AUC, where the performance of the model is evaluated using the averaged prediction between CC and MLO views of each breast. 
    This is the main evaluation metric of this work.
\end{enumerate}

To address the class imbalance of the dataset, we additionally report specificity and Matthew's correlation coefficient~\cite{matthews1975comparison} at 90\% sensitivity. \footnote{If no threshold leads to exactly 90\% sensitivity, then the threshold that leads to the closest sensitivity value is chosen. When calculating confidence intervals with bootstrapping, trials for which the difference between the ideal and actual sensitivity is greater than 2.5 percentage points are discarded.}

\subsection{Semantic segmentation}
We calculate the Dice similarity coefficient (DSC) and pixel average precision (PxAP) \cite{choe2020evaluating} to evaluate the ability of 3D-GMIC to perform weakly-supervised semantic segmentation based on the saliency maps.
DSC and PxAP are computed as an average over the images with visible biopsy-confirmed findings. 
It is worth noting that while comparing these values to different models learning from the same imaging modality is fair, comparing them between models learning from different imaging modalities is not necessarily so.
FFDM images have higher resolutions than both DBT and synthetic 2D images, and therefore each superpixel in the saliency maps corresponds to a smaller physical region.
This results in finer saliency maps for FFDM images.
In addition, DBT has a lower ratio of important regions compared to the entire image because the lesions only appear in a small fraction of slices in each image.
Therefore, DSC or PxAP values will be lower for DBT even if the model similarly highlights important regions in the slices where they are visible.
To partially mitigate this issue and make these values more comparable between 2D and 3D imaging modalities, we do not report the DSC or PxAP on DBT images and instead calculate these metrics after max-projecting both saliency maps and the segmentation labels in the depth dimension.
This way, we can get a better sense of how 3D-GMIC is predicting the location of lesions in the xy-dimension by comparing its score to the 2D models.

\section{Results}

We evaluated the classification and semantic segmentation performances of GMIC and 3D-GMIC architectures on the internal NYU Combo v1 dataset as well as an external DBT dataset from Duke University hospital.

\subsection{Computational efficiency}


Compared to off-the-shelf deep convolutional neural networks, 3D-GMIC uses 77.98\%-90.05\% less GPU memory and 91.23\%-96.02\% less computation as shown in Table~\ref{tb:computational_efficiency}.

\begin{table}[htb!]
    \centering
      \caption{Computational efficiency of processing a DBT image with 96 slices. 3D-GMIC uses 77.98\%-90.05\% less GPU memory and 91.23\%-96.02\% less computation than other architectures.}
\begin{tabular}{l|rr}
\hline 
 & Maximum GPU RAM usage (MB) & GMACs \\ \hline \hline
\shortstack[c]{ResNet-18~\cite{he2015deepresidual}}                &   \shortstack[c]{171,031}  & \shortstack[c]{9,932}   \\
\shortstack[c]{ResNet-34~\cite{he2015deepresidual}}           &   \shortstack[c]{223,235}  & \shortstack[c]{20,056}   \\
\shortstack[c]{3D ResNet-18~\cite{hara3dcnns}}                 &   \shortstack[c]{100,903}  & \shortstack[c]{9,104} \\ 
\shortstack[c]{3D ResNet-34~\cite{hara3dcnns}}                &   \shortstack[c]{111,650}  & \shortstack[c]{12,192}   \\
\shortstack[c]{3D-GMIC}           &   \shortstack[c]{\textbf{22,219}}  & \shortstack[c]{\textbf{798}}   \\ \hline
\end{tabular}
  \label{tb:computational_efficiency}
\end{table}

\begin{table*}[ht]
    \centering
    \caption{Breast-wise test performances (AUC) for the global-only and full architectures. 
    For both global-only and full architectures, we find that 3D-GMIC performs comparably with other models trained with 2D imaging modalities. In addition, a multi-modal ensemble including all three modalities leads to the best performance in predicting the presence of malignant lesions. We compute 95\% confidence intervals using 1,000 iterations of the bootstrap method~\cite{johnson2001introduction}.}
    \begin{tabular}{lcccc}
\hline
    & \multicolumn{2}{c}{global-only}   & \multicolumn{2}{c}{full}  \\ \cline{2-5}
      Modality & Malignant       & Benign        & Malignant      & Benign    \\ \hline \hline
\shortstack[c]{FFDM}      & \shortstack[c]{0.802 (0.734-0.864)}           & \shortstack[c]{0.710 (0.681-0.739)}          & \shortstack[c]{0.816 (0.737-0.878)}          & \shortstack[c]{\textbf{0.728} (0.695-0.758)}  \\ 
\shortstack[c]{Synthetic 2D}  & \shortstack[c]{0.790 (0.719-0.856)}           & \shortstack[c]{0.696 (0.662-0.727)}          & \shortstack[c]{0.826 (0.754-0.884)}          & \shortstack[c]{0.699 (0.666-0.732)}          \\ 
\shortstack[c]{DBT}      & \shortstack[c]{\textbf{0.811} (0.747-0.869)}   & \shortstack[c]{\textbf{0.714} (0.683-0.747)} & \shortstack[c]{\textbf{0.831} (0.769-0.887)} & \shortstack[c]{0.717 (0.687-0.745)}         \\ \cline{1-5}
\shortstack[c]{FFDM + Synthetic 2D}    & \shortstack[c]{0.805 (0.734-0.873)}  & \shortstack[c]{0.709 (0.679-0.737)}           & \shortstack[c]{0.832 (0.757-0.892)}          & \shortstack[c]{0.720 (0.689-0.751)}        \\ 
\shortstack[c]{DBT + Synthetic 2D}     & \shortstack[c]{0.807 (0.733-0.871)}           & \shortstack[c]{0.712 (0.682-0.744)}           & \shortstack[c]{0.840 (0.775-0.894)}          & \shortstack[c]{0.715 (0.682-0.746)}         \\ 
\shortstack[c]{DBT + FFDM}          & \shortstack[c]{\textbf{0.812} (0.743-0.871)}           & \shortstack[c]{\textbf{0.721} (0.691-0.749)}  & \shortstack[c]{0.837 (0.767-0.892)}          & \shortstack[c]{\textbf{0.728} (0.695-0.759)}   \\ 
\shortstack[c]{All 3 modalities}          & \shortstack[c]{0.811 (0.739-0.875)}           & \shortstack[c]{0.717 (0.687-0.745)}           & \shortstack[c]{\textbf{0.841} (0.768-0.895)} & \shortstack[c]{0.723 (0.692-0.754)}        \\ \hline 
\end{tabular}
    \label{tab:auc_table}
\end{table*}

\begin{table*}[ht]
    \centering
    \caption{Breast-wise test performances (specificity at 90\% sensitivity) for the global-only and full architectures. 
    We compute 95\% confidence intervals using 1,000 iterations of the bootstrap method~\cite{johnson2001introduction}.}
    \begin{tabular}{lcccc}
\hline
    & \multicolumn{2}{c}{global-only}   & \multicolumn{2}{c}{full}  \\ \cline{2-5}
      Modality & Malignant       & Benign        & Malignant      & Benign    \\ \hline \hline
\shortstack[c]{FFDM} & \shortstack[c]{0.401 (0.146-0.689)} & \shortstack[c]{\textbf{0.365} (0.282-0.439)} & \shortstack[c]{0.448 (0.120-0.716)} & \shortstack[c]{0.348 (0.262-0.436)} \\ 
\shortstack[c]{Synthetic 2D} & \shortstack[c]{0.450 (0.089-0.610)} & \shortstack[c]{\textbf{0.365} (0.258-0.418)} & \shortstack[c]{0.474 (0.162-0.673)} & \shortstack[c]{0.345 (0.250-0.411)} \\ 
\shortstack[c]{DBT} & \shortstack[c]{\textbf{0.503} (0.218-0.666)} & \shortstack[c]{0.349 (0.312-0.444)} & \shortstack[c]{\textbf{0.554} (0.249-0.703)} & \shortstack[c]{\textbf{0.383} (0.304-0.437)} \\ \hline 
\shortstack[c]{FFDM + Synthetic 2D} & \shortstack[c]{0.427 (0.138-0.667)} & \shortstack[c]{0.379 (0.271-0.472)} & \shortstack[c]{0.475 (0.074-0.724)} & \shortstack[c]{\textbf{0.413} (0.259-0.458)} \\ 
\shortstack[c]{DBT + Synthetic 2D } & \shortstack[c]{\textbf{0.489} (0.123-0.629)} & \shortstack[c]{0.389 (0.314-0.438)} & \shortstack[c]{\textbf{0.597} (0.181-0.713)} & \shortstack[c]{0.378 (0.304-0.437)} \\ 
\shortstack[c]{DBT + FFDM 2D} & \shortstack[c]{0.417 (0.179-0.707)} & \shortstack[c]{0.374 (0.312-0.434)} & \shortstack[c]{0.585 (0.197-0.729)} & \shortstack[c]{0.379 (0.288-0.435)} \\ 
\shortstack[c]{All 3 modalities} & \shortstack[c]{0.466 (0.126-0.692)} & \shortstack[c]{\textbf{0.399} (0.313-0.446)} & \shortstack[c]{0.553 (0.141-0.744)} & \shortstack[c]{0.400 (0.293-0.446)} \\ \hline 
\end{tabular}
    \label{tab:specificity_table}
\end{table*}

\begin{table*}[ht]
    \centering
    \caption{Breast-wise test performances (Matthew's correlation coefficient at 90\% sensitivity) for the global-only and full architectures. 
    We compute 95\% confidence intervals using 1,000 iterations of the bootstrap method~\cite{johnson2001introduction}.}
    \begin{tabular}{lcccc}
\hline
    & \multicolumn{2}{c}{global-only}   & \multicolumn{2}{c}{full}  \\ \cline{2-5}
      Modality & Malignant       & Benign        & Malignant      & Benign    \\ \hline \hline
\shortstack[c]{FFDM} & \shortstack[c]{0.031 (0.008-0.064)} & \shortstack[c]{\textbf{0.057} (0.042-0.072)} & \shortstack[c]{0.035 (0.001-0.067)} & \shortstack[c]{0.054 (0.038-0.071)} \\ 
\shortstack[c]{Synthetic 2D} & \shortstack[c]{0.035 (-0.005-0.052)} & \shortstack[c]{\textbf{0.057} (0.037-0.068)} & \shortstack[c]{0.037 (0.006-0.063)} & \shortstack[c]{0.054 (0.036-0.067)} \\ 
\shortstack[c]{DBT} & \shortstack[c]{\textbf{0.040} (0.011-0.061)} & \shortstack[c]{0.054 (0.046-0.073)} & \shortstack[c]{\textbf{0.046} (0.015-0.068)} & \shortstack[c]{\textbf{0.061} (0.045-0.072)} \\  \hline 
\shortstack[c]{FFDM + Synthetic 2D} & \shortstack[c]{0.033 (0.003-0.060)} & \shortstack[c]{0.060 (0.040-0.078)} & \shortstack[c]{0.038 (-0.005-0.070)} & \shortstack[c]{\textbf{0.066} (0.037-0.076)} \\ 
\shortstack[c]{DBT + Synthetic 2D} & \shortstack[c]{\textbf{0.039} (0.001-0.054)} & \shortstack[c]{0.062 (0.047-0.072)} & \shortstack[c]{\textbf{0.051} (0.009-0.068)} & \shortstack[c]{0.060 (0.046-0.072)} \\ 
\shortstack[c]{DBT + FFDM} & \shortstack[c]{0.032 (0.008-0.068)} & \shortstack[c]{0.059 (0.047-0.070)} & \shortstack[c]{0.049 (0.010-0.071)} & \shortstack[c]{0.060 (0.043-0.071)} \\ 
\shortstack[c]{All 3 modalities} & \shortstack[c]{0.037 (0.002-0.063)} & \shortstack[c]{\textbf{0.064} (0.048-0.074)} & \shortstack[c]{0.045 (0.004-0.072)} & \shortstack[c]{0.064 (0.045-0.074)} \\ \hline 
\end{tabular}
    \label{tab:mcc_table}
\end{table*}

\subsection{Classification}

For each imaging modality, we trained the GMIC or 3D-GMIC architecture as well as a variant with only \textit{the global module} (global-only).
This measures the benefit of adding \textit{the local module} in the GMIC architecture.
The results are in Table~\ref{tab:auc_table}, Table~\ref{tab:specificity_table}, and Table~\ref{tab:mcc_table}.
We also show the ensemble performances where the predictions of different models, trained with different modalities and/or for a different number of epochs are averaged.

For both global-only and full architectures, we find that 3D-GMIC performs comparably with other models trained with 2D imaging modalities.
Ensembling models trained with different modalities does help, but which modalities lead to the best performance differs between models and target tasks.

In addition, we demonstrate our models' classification performances on the BCS-DBT dataset from Duke University Hospital \cite{buda2021data, buda2020data, clark2013cancer}.
3D-GMIC achieved image-wise AUC of 0.848 (95\% CI: 0.798-0.896) in identifying images with malignant findings and image-wise AUC of 0.741 (95\% CI: 0.697-0.785) in identifying images with benign findings. 
In comparison, on the DBT images in the test set of NYU Combo v1 dataset, the same 3D-GMIC models achieve image-wise AUC of 0.809 (95\% CI: 0.761-0.852) in identifying images with malignant findings and image-wise AUC of 0.704 (95\% CI: 0.683-0.728) in identifying images with benign findings.
Even though the model did not use any images from Duke during training, it generalizes well and even shows higher performances than on our own internal data set.

\subsection{Semantic segmentation}

\begin{figure*} 
    \centering
    \vspace{-4mm}
\includegraphics[width=0.9\textwidth]{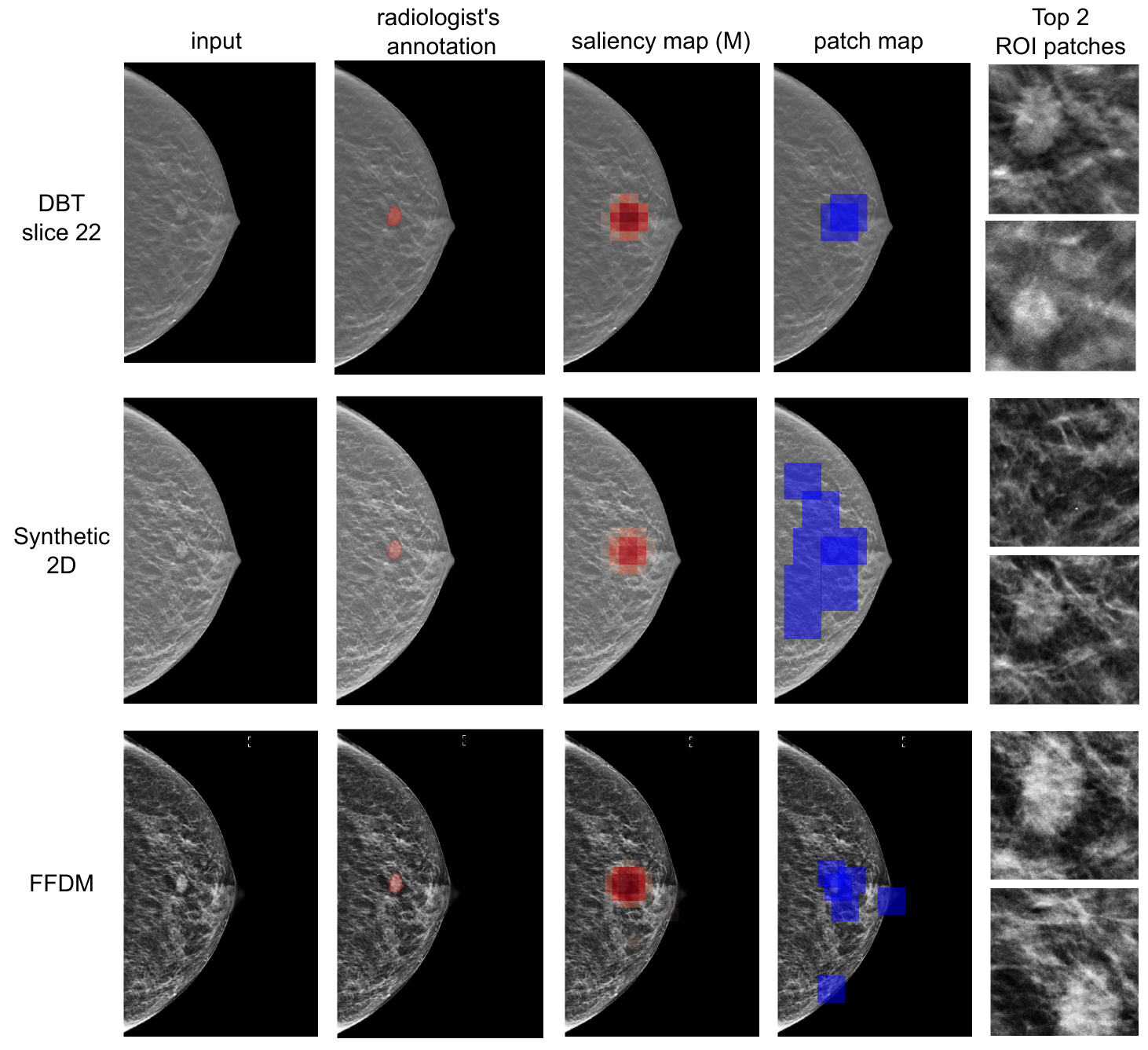}
\vspace{-1mm}
    \caption{
    Visualization of results for an example for DBT, synthetic 2D and FFDM modalities. 
    From left to right: input images, patch maps indicating the locations of cropped ROI patches as blue squares, saliency maps for benign category, saliency maps for malignant category, and the cropped patches with the two highest attention scores.
    This example contains an irregular mass in the central right breast at mid depth was diagnosed as malignant on ultrasound-guided core biopsy.
    }
  \label{fig:vis_plot_1}
\vspace{-3mm}
\end{figure*}

Table~\ref{tab:localization} shows DSC and PxAP scores for the three modalities.
Even though semantic segmentation in 3D data can be more difficult, 3D-GMIC reaches comparable performances. 
%
A sample set of model visualizations, produced for the same breast, is shown in Fig.~\ref{fig:vis_plot_1}.

\begin{table}[ht]
    \centering
    \caption{Weakly-supervised semantic segmentation performances on the test set. The mean and standard deviation of individual performances of the five best models and their ensemble performances. ``DBT max-projected'' refers to the evaluation where both saliency maps and the annotations were max-projected in the depth dimension. 3D-GMIC reaches comparable DSC and PxAP scores.}
    \resizebox{\columnwidth}{!}{

\begin{tabular}{l|cc|cc}
\hline
       & DSC(M)          & DSC(B)          & PxAP(M)         & PxAP(B)         \\ \hline \hline
FFDM        & \textbf{0.193} ± 0.019 & 0.120 ± 0.025 & \textbf{0.100} ± 0.016 & 0.049 ± 0.013 \\ 
Synthetic 2D & 0.166 ± 0.025 & \textbf{0.144} ± 0.023 & 0.092 ± 0.016 & \textbf{0.070} ± 0.009 \\
DBT max-projected & 0.150 ± 0.018 & 0.123 ± 0.017 & 0.065 ± 0.008 & 0.055 ± 0.009 \\ \hline 
FFDM ensemble       & \textbf{0.208} & 0.138 & \textbf{0.111} & 0.053 \\ 
Synthetic 2D ensemble& 0.181 & \textbf{0.162} & 0.101 & \textbf{0.082} \\
DBT max-projected ensemble & 0.162 & 0.141 & 0.066 & 0.060 \\ \hline 
\end{tabular}
}
    \label{tab:localization}
\end{table}

\subsection{Ablation studies}
We performed several ablation studies to assess the effects of different hyperparameters on 3D-GMIC. 

\subsubsection{Comparing modalities using the same number of patches}
Table~\ref{tb:ablation_numpatches} shows the comparison of model performances when using the same number of patches. 
We observe that 3D-GMIC on DBT still shows comparable performances to GMIC on 2D images.

\begin{table}[htb!]
    \centering
      \caption{Ablation study: the breast-wise test performances of models trained with each modality when cropping 8 ROI patches in $\texttt{retrieve\_roi\_from\_3d\_image}$ algorithm. We train three models for each setting and show ensemble performances with ten random augmentations, along with the 95\% confidence intervals. 
      3D-GMIC on DBT is still showing comparable performances to GMIC on 2D images, even when the models use the same number of patches.
      }
\begin{tabular}{l|cc}
\hline 
 & AUC (M) & AUC (B) \\ \hline \hline
\shortstack[c]{FFDM}                &   \shortstack[c]{0.796 (0.719-0.860)}  & \shortstack[c]{0.707 (0.676-0.740)}   \\
\shortstack[c]{Synthetic 2D}           &   \shortstack[c]{0.821 (0.744-0.886)}  & \shortstack[c]{0.705 (0.672-0.735)}   \\
\shortstack[c]{DBT}                 &   \shortstack[c]{\textbf{0.834} (0.773-0.888)}  & \shortstack[c]{\textbf{0.717} (0.686-0.749)} \\ \hline
\end{tabular}
  \label{tb:ablation_numpatches}
\end{table}

\subsubsection{The half-width of slice-wise sampling $\zeta$}

The default value of $\zeta=10$ was set in initial experiments before systematically tuning the hyperparameters and was used in our experiments. 
In this ablation study, we investigate the effect of the half-width of slice-wise sampling $\zeta \in \{0, 5, 10, \infty\}$ during training by keeping the rest of the hyperparameters the same. 
The purpose of this ablation study is twofold.
First, we aim to observe if slice-wise patch sampling provides an unfair advantage to 3D-GMIC compared to other GMIC models trained on 2D modalities.
Second, we want to verify if the default value of $\zeta=10$ was optimal.
The results are in Table~\ref{tb:topk_tb}.
We show that the performance of $\zeta=0$ and $\zeta=10$ are indistinguishable, which means that the slice-wise sampling of the image patches did not give an unfair advantage to 3D-GMIC compared to GMIC with 2D imaging modalities.
In addition, we observe that $\zeta=5$ leads to an improved performance.

\begin{table}[htb!]
    \centering
      \caption{Ablation study: effect of a choice of $\zeta$ on the AUC of identifying breasts with malignant findings, along with the 95\% confidence intervals.
      }
    \begin{tabular}{l | c  }
        \hline  
                 & AUC (M) \\ \hline \hline 
        \shortstack[c]{$\zeta$=0} & \shortstack[c]{0.830 (0.774-0.880)} \\
        \shortstack[c]{$\zeta$=5} & \shortstack[c]{\textbf{0.852} (0.789-0.901)} \\
        \shortstack[c]{$\zeta$=10} & \shortstack[c]{0.834 (0.773-0.888)}   \\
        \shortstack[c]{$\zeta$=$\infty$} & \shortstack[c]{0.835 (0.770-0.895)} \\ \hline
    \end{tabular}
  \label{tb:topk_tb}
\end{table}

\section{Discussion and Conclusion}
In this work, we propose 3D-GMIC, a novel deep neural network capable of learning and predicting from high-resolution 3D medical images in a computationally efficient manner.
3D-GMIC effectively focuses its computation to the small subset of important regions by first identifying the regions of interest with a low-capacity sub-network and selectively applying a high-capacity sub-network to the regions of interest while avoiding processing duplicate information from nearby slices.
3D-GMIC enables training deep neural networks for high-resolution 3D images with small regions of interests without requiring expensive annotation labels or compromising performance. 
Our model is efficient enough to train with a batch size of 4 images using 4 GPUs with 32GB of GPU memory.

3D-GMIC focuses its computation on a lower proportion of the input image compared to GMIC.
This is done by avoiding cropping patches with duplicate information as explained in section~\ref{sec:local}. 
For example, consider GMIC on synthetic 2D image of size (2116 x 1339) and 3D-GMIC on DBT image of size (2116 x 1339 x 70), both cropping 8 two-dimensional patches of size 256x256 as shown in Table~\ref{tb:ablation_numpatches}. 
Then the local networks of either model will utilize 256*256*8=524,288 pixels. 
While this corresponds to 18.50\% of the synthetic 2D image, it corresponds to only 0.26\% of the DBT image. 

We demonstrate the performance of the proposed architecture with a large dataset where each exam contains the three imaging modalities of screening mammography: FFDM, synthetic 2D, and DBT.
3D-GMIC achieved comparable performances compmared to GMIC on FFDM or synthetic 2D.
This demonstrates that 3D-GMIC successfully classified large 3D images despite focusing its computation on a smaller percentage of its input compared to GMIC.
Furthermore, the performance is improved when we ensemble predictions from multiple imaging modalities.
This suggests that the model might have learned different behaviors for each modality depending on their relative strengths and weaknesses.
This accentuates the benefit of training AI systems which can handle the DBT images even when the 2D imaging modalities with equivalent information are available.


The reported semantic segmentation and classification performances of 2D modalities are lower than the values reported in the original GMIC paper.
The reason for this is twofold.
First, the NYU Combo v1 dataset is smaller than the BCSDv1 dataset used in the original GMIC paper. 
Our work utilizes only half the number of exams with biopsy labels and this leads to overfitting more quickly.
Second, the two papers have different test sets and two AUCs calculated from different datasets are not directly comparable.


We note that 3D-GMIC shares some limitations with the standard GMIC model. Namely, training 3D-GMIC is more complicated than training 3D ResNet models because there are two separate networks to optimize. The learning speeds for the global and local modules could be different and they could start overfitting at different epochs. This could prevent 3D-GMIC from reaching the best possible performance. To mitigate this, separate learning rates could be used for global and local modules as done in Wu et al. (2020)~\cite{wu2020improving}. 


\bibliography{IEEEabrv.bib,ref.bib}{}
\bibliographystyle{IEEEtran}

\end{document}